\documentclass{article}

\usepackage[final]{neurips_2024}

\usepackage[utf8]{inputenc} 
\usepackage[T1]{fontenc}    
\usepackage{hyperref}       
\usepackage{url}            
\usepackage{booktabs}       
\usepackage{amsfonts}       
\usepackage{nicefrac}       
\usepackage{microtype}      
\usepackage{xcolor}         
\usepackage{wrapfig}

\usepackage{algorithm}
\usepackage{algorithmic}

\usepackage{booktabs} 
\usepackage{multirow}
\usepackage{array}
\usepackage{amsmath}
\usepackage{bbold}
\usepackage{svg}

\usepackage{xcolor}  

\definecolor{soham}{rgb}{0.8, 0.2, 0.2}

\definecolor{blue}{rgb}{0.0, 0.5, 1.0}  

\definecolor{q}{rgb}{0.09, 0.61, 0.1} 

\definecolor{m}{rgb}{0.6, 0.33, 0.73}

\newcommand{\bigO}{\mathcal{O}}
 
\newcommand{\hmme}{\text{HMM-\textit{e}\ }}

\usepackage{subfiles}
\setcitestyle{numbers}

\title{Behavioral Sequence Modeling \\with Ensemble Learning}

\author{%
  Maxime Kawawa-Beaudan \\
  J.P. Morgan AI Research \\
  383 Madison Avenue, New York NY USA\\
  \texttt{maxime.kawawa-beaudan@jpmorgan.com} \\
  \And
  Srijan Sood \\
  J.P. Morgan AI Research \\
  383 Madison Avenue, New York NY USA\\
  \texttt{srijan.sood@jpmorgan.com} \\
  \And
  Soham Palande \\
  J.P. Morgan AI Research \\
  383 Madison Avenue, New York NY USA\\
  \texttt{soham.palande@jpmorgan.com} \\
  \And
  Ganapathy Mani \\
  J.P. Morgan AI Research \\
  383 Madison Avenue, New York NY USA\\
  \texttt{ganapathy.mani@jpmorgan.com} \\
  \And
  Tucker Balch \\
  Emory University \\
  1300 Clifton Road NE, Atlanta GA USA\\
  \texttt{tucker.balch@emory.edu} \\
  \And
  Manuela Veloso \\
  J.P. Morgan AI Research \\
  383 Madison Avenue, New York NY USA\\
  \texttt{manuela.veloso@jpmorgan.com} \\
}

\begin{document}

\maketitle

\begin{abstract}
We investigate the use of sequence analysis for behavior modeling, emphasizing that sequential context often outweighs the value of aggregate features in understanding human behavior. We discuss framing common problems in fields like healthcare, finance, and e-commerce as sequence modeling tasks, and address challenges related to constructing coherent sequences from fragmented data and disentangling complex behavior patterns. We present a framework for sequence modeling using Ensembles of Hidden Markov Models, which are lightweight, interpretable, and efficient. Our ensemble-based scoring method enables robust comparison across sequences of different lengths and enhances performance in scenarios with imbalanced or scarce data. The framework scales in real-world scenarios, is compatible with downstream feature-based modeling, and is applicable in both supervised and unsupervised learning settings. We demonstrate the effectiveness of our method with results on a longitudinal human behavior dataset. 

\end{abstract}

\section{Introduction} \label{sec:introduction}

Modeling human behavior is a complex task with applications spanning multiple domains such as user research, healthcare, payments, trading, and e-commerce. Applications range from classifying human activity \cite{har_video}, distinguishing humans from bots \cite{bot_detection_dl}, detecting credit card fraud \cite{hmm_card_fraud} etc. 

Behavior is often captured as sequences of actions or events over time, and understanding patterns within these sequences is crucial for tasks like classification, anomaly detection, and user modeling. A key challenge in utilizing the captured sequences is class imbalance, where underrepresented behavior profiles or a disproportionate number of anomalous examples hinder model generalization.

Many solutions leverage complex deep learning models~\cite{yu2024creditcardfrauddetection}, focusing on event-level classification with extensive feature engineering~\cite{feature_engineering_credit_card_fraud}. However, such event-level or feature-aggregated methods frequently fall short in capturing the sequential dynamics essential for understanding and modeling behavior. These approaches are also susceptible to overfitting, with performance rapidly degrading in class-imbalanced scenarios.

Sequential context is crucial for effective behavior modeling. For example, in credit card fraud detection or anti-money laundering, a user’s transaction history provides deeper insights into behavioral intent than isolated transactions or aggregated features. Yet, many datasets and approaches remain confined to the event level, overlooking the broader sequential context.
In this study, we advocate for a sequence modeling approach to behavior modeling, particularly in scenarios where behaviors are represented as action sequences derived from unstructured data. Aggregating this data into coherent sequences that reflect an agent’s decision-making process is a non-trivial challenge. Our method is tested in a real-world setting characterized by extreme class imbalance, with millions of diverse user behavior examples contrasted against only a few hundred instances of the anomalous class.

We propose a novel and lightweight ensemble-based framework for behavior modeling, and show efficacy on downstream (imbalanced) sequence classification tasks. While our approach is model-agnostic, we leverage Hidden Markov Models (HMMs) for their simplicity, interpratibility, and efficacy at capturing temporal dependencies and latent patterns. Our real-world deployment scales to millions of sequences, while being compatible with downstream machine learning methods. We demonstrate its effectiveness on a publicly available human behavior dataset \cite{globem}. 

Our contributions: (1) We introduce a behavior modeling framework based on sequences of events/actions, applicable to various domains; (2) We outline its application to supervised and unsupervised tasks; (3) We demonstrate its effectiveness on a longitudinal human behavior dataset.

\section{Background \& Related Work} \label{sec:background}

\paragraph{HMMs} Hidden Markov Models (HMMs) are statistical models for sequential data, which have a long history of use in natural language processing, finance, and bioinformatics \cite{hmm_intro_1986, hmm_biology_recent_survey, hmm_genomics_survey, hmm_stock_selection}. HMMs have been used extensively for behavior modeling, including sensor surveillance \cite{hmm_surveillance}, human-computer interfaces \cite{hmm_headmouse}, and web user interactions \cite{hmm_web_user}, with recent applications in social media bot detection~\cite{bot_detection_hmm_recent}. While neural network-based approaches like CNNs, LSTMs, and Transformers have shown success in settings like sentiment analysis \cite{kansara2020traditionalvsdeeplearning} and network intrusion detection \cite{network_detection_transformers_imbalanced}, they face challenges such as high computational cost, overfitting, and reduced interpretability.

\paragraph{Resolution}

Event-level classification still dominates in areas like anti-money laundering and network security, where sequence-level labels are often missing \cite{aml_synthetic, network_intrusion_dataset_overview}. This lends itself to aggregate feature based approaches, missing key historical context. While some work has tackled sequence modeling in network intrusion detection with favorable results \cite{dl_network_sequential}, much remains to be done.

\paragraph{Data Imbalance and Anomaly Detection}

Many real-world problems, including intrusion detection \cite{intrusion_detection_review}, credit card fraud \cite{hmm_card_fraud}, and money laundering \cite{jpm_air_aml}, involve detecting rare events and suffer from class imbalance. One-class anomaly detection focuses on robustly modeling the nominal class and identifying deviations \cite{nominal_behavior_modeling_autoencoders}, while more targeted approaches model both normal and anomalous sequences to detect specific behavioral anomalies \cite{imbalanced_sequence_classification}.

\section{Approach}

Consider a sequence observation $\bigO = \{a_1, a_2, ..., a_T\}$, where each $a_i$ is drawn from a discrete set of actions $\mathcal{A}$. Such sequences can represent various behaviors, such as user interactions in an app, trading actions in financial markets, or other human decision-making processes. Our goal is to model these behaviors, either discovering behavior clusters, or classifying behaviors when labels are available (e.g., online bot detection, credit card fraud detection, physical activity recognition~\cite{botdetection_sequencemodeling, yu2024creditcardfrauddetection, human_activity_recognition_spikenn}.

\subsection{Sequence Construction} \label{sec:sequence_construction}

One of the primary challenges lies not in modelling but in organizing coherent data streams from raw, fragmented data $\mathcal{D}$, which often contains interwoven behaviors from multiple agents/users. 

For instance, in trading, $\mathcal{D}$ may span billions of transactions across participants, assets, and exchanges, requiring grouping data streams by participant, and further by exchange or asset to capture specific behaviors. In network analysis, interactions between  devices and servers can be grouped by source IP for individual user activity, or further by target IP to constitute specific behavior streams.

\begin{figure*}[tb]
\centering
\includegraphics[width=1.0\linewidth]{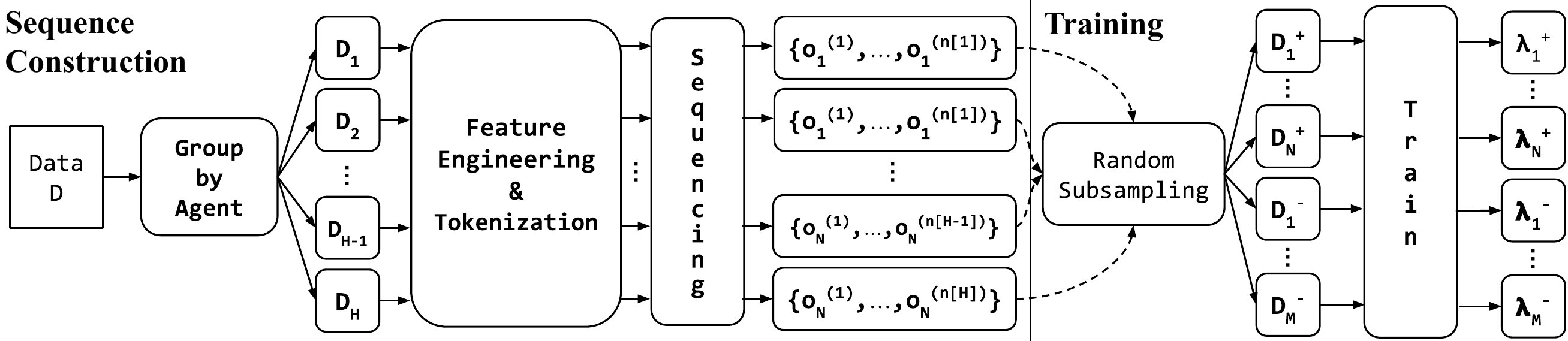} 
\caption{Flow diagram of our sequence construction approach, as detailed in Section \ref{sec:sequence_construction}. We disentangle the monolithic dataset $\mathcal{D}$ into data streams, then process these further into sets of observation sequences. Our subsequent \hmme ensemble training approach is detailed in Section \ref{sec:hmm_ensembles}. While we adopt this approach using HMMs, the framework itself is model agnostic. The training data is broken into random subsets, and a diverse ensemble of learners is trained on these subsets.}
\label{fig:data_approach_diagram}
\end{figure*}

As illustrated in Fig.~\ref{fig:data_approach_diagram}, we begin by disentangling $\mathcal{D}$ into separate data streams $\mathcal{D}_1, ..., \mathcal{D}_H$, each corresponding to one of  H  agents. Feature engineering refines these data streams through dimensionality reduction, tokenization, or discretization, enhancing model generalization, particularly in the presence of imbalanced or sparse datasets.Continuous features can also be normalized and estimated directly, through techniques like Gaussian HMMs~\cite{gaussian_hmm_human_activity}.

Once data is organized into streams $\mathcal{D}_h$, sequences of observations ${\bigO_h^{(1)}, ..., \bigO_h^{(n[h])}}$ are then extracted from each stream, with domain knowledge or sessions guiding the sequence span (start and end points). For example, web user behavior may span minutes to hours, whereas medical trial observations could extend over days or weeks. Breaks in continuous data streams often demarcate sequences, with shorter pauses treated as wait events and longer breaks as sequence endpoints. The number of sequences can vary significantly across agents, reflecting differing activity levels (e.g. power users vs intermittent monthly users).

\subsection{Ensembles of Hidden Markov Models}

\textbf{Singleton HMMs \hspace{0.5em}}
First, considering binary sequence classification, we separate training data by class and train two indivudal HMMs: one positive class HMM $\lambda^+$, and one negative class HMM $\lambda^-$. Given an unseen sequence $\bigO$, the predicted class $c(\bigO)$ is determined by comparing the likelihoods: 

\begin{equation}
\label{eqn:max_likelihood_classifier}
c(\bigO) = \mathbb{1}\{p(\bigO \mid \lambda^+) > p(\bigO \mid \lambda^-)\}
\end{equation}

\textbf{Variable Sequence Length \hspace{0.5em}} \label{subsection_variable_sequence_length}
HMMs excel at sequence analysis~\cite{hmm_biology_recent_survey}, but struggle when comparing sequences of varying length, as length influences likelihood computation exponentially. We address this through model-driven normalization, computing likelihoods for a given sequence across multiple models, and deriving a rank-based composite score (rather than comparing sequence likelhihoods).

\textbf{HMM Ensembles \hspace{0.5em}}\label{sec:hmm_ensembles}
HMMs, while lightweight and efficient, can struggle to capture the complexity of behaviors in training data when using a singular model (per class). Ensemble methods train multiple models on subsets of the data \cite{dietterich2000ensemble}, enabling each learner to specialize on distinct patterns or behaviors, while collectively capturing the full data distribution. This results in a more robust approach, particularly in scenarios with data imbalance, where monolithic models skew towards modeling the majority class (or underfitting for class specific models) \cite{kuncheva2014combining}.
We propose \textbf{HMM-\textit{e}}, an ensemble framework that computes composite scores from individual learners~\cite{hmme_preprint}. While HMMs are effective for our case, this framework is model-agnostic and can incorporate other model classes such as neural networks, SVMs, or decision trees.

\subsubsection{Formalization and Algorithmic Framework}\label{sec:formalization}
First, we train $N$ models $\{\lambda_1^+,...,\lambda_N^+\}$ on the positive class and $M$ models $\{\lambda_1^-,...,\lambda_M^-\}$ on the negative class, taking care to ensure diversity among the models by training each on a randomly selected subset of samples from the training data. Each model sees $s\%$ of the training data in its relevant class. While we set $N = M$ for all settings, these parameters ($M, N, s$) can be established using typical hyperparameter optimization approaches. For any given sequence in the training data, the probability of not being selected for any model's random subset is $(1-s)^N$. The expected number of unsampled sequences is the same, so it is important to select $s$ and $N$ to keep this proportion of the training data small. 

For an unseen observation sequence $\bigO$, we compute its likelihood scores under all models: $\{p(\bigO \mid \lambda_1^+),...,p(\bigO \mid \lambda_N^+)\}$ and $\{p(\bigO \mid \lambda_1^-),...,p(\bigO \mid \lambda_M^-)\}$. We then compute a composite score: 

\begin{equation}
\label{eqn:composite_score}
s(\bigO) = \sum_{i=1}^N \sum_{j=1}^M \mathbb{1} \{p(\bigO \mid \lambda_i^+) > p(\bigO \mid \lambda_j^-) \}  
\end{equation}

The score $s(\bigO)$ represents the pairwise comparisons where positive-class models assign a higher likelihood than negative-class models, taking values in $[0, N \times M]$. A low score indicates that the sequence is more likely under the negative-class models, and vice versa.
As likelihoods across different sequence lengths are not directly compared, this composite score acts as an implicit normalization technique. $N$ and $M$ should be chosen such that the score range adequately distinguishes the classes.

For our HMM and \hmme approaches, we use 3 states in each of our models, and converge on an ensemble size of 250 and a subset factor of 1\%. We perform hyperparameter search for ensemble size, trying other values in [10, 50, 100, 500, 1000]. We settle on 250 for its good performance at a relatively low complexity.

\textbf{Downstream Modeling using \hmme Scores \hspace{0.5em}} \label{sec:downstream_modeling}
Given a corpus of sequences and corresponding scores $\{\bigO, s(\bigO)\}$, we classify sequences using a threshold $s_{thresh}$:  $c(\bigO_i) = \mathbb{1}\{ s(\bigO_i) \geq s_{thresh} \}$. Alternatively, base learner likelihoods can serve as features for downstream classifiers. For each sequence $\bigO_i$, we define a feature vector
\begin{equation}
        f_i = \begin{bmatrix}
           p(\bigO_i | \lambda_1^+),
           \hdots, 
           p(\bigO_i | \lambda_N^+),
          p(\bigO_i | \lambda_1^-),
           \hdots,
           p(\bigO_i | \lambda_M^-)
         \end{bmatrix}
\end{equation}
To account for sequence length sensitivity,  $f_i$ is normalized by $\lvert\lvert  f_i \rvert\rvert_2$.
This technique utilizes HMMs as feature extractors, where each feature $p(\bigO_i | \lambda_j)$ represents the similarity between the sequence $\bigO_i$, and the random subset of training data underlying $\lambda_j$.

\begin{wrapfigure}[18]{r}{0.4\textwidth}
\centering
\includegraphics[width=0.4\textwidth]{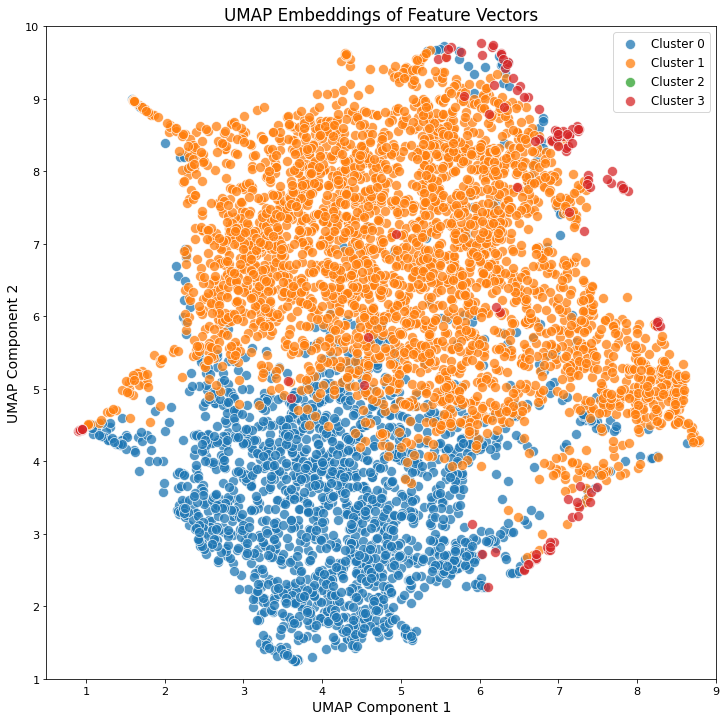} \caption{UMAP embeddings of features $f_i$, as discussed in Section \ref{sec:clustering_unsupervised}, from a 500-model ensemble. Colors correspond to clusters discovered via K-Means.}
\label{fig:umap_unsupervised}
\end{wrapfigure}

\textbf{Clustering \hmme Scores in Unsupervised Settings \hspace{0.5em}} \label{sec:clustering_unsupervised}
In label-free settings, behavior clustering can be achieved using unsupervised learning approaches. We train $N$ models $\{\lambda_1,...,\lambda_N\}$ on random $s\%$ data subsets and generate feature vectors of base learner likelihoods $f_i$ (Section~\ref{sec:formalization}). Unsupervised clustering like K-Means can be applied to discover behavioral groups, dimensionality reduction (e.g., PCA) can be helpful when $N$ is large.

\section{Experiments}

\textbf{Data \hspace{0.5em}} \label{sec:data}

We evaluate our approach on the GLOBEM dataset \cite{globem}, a longitudinal human behavior study featuring over 3,700 attributes from 497 participants across four years (2018-2021). The dataset includes survey, smartphone, and wearable data, with a focus on depression detection (the task we consider). Data includes mood assessments, step counts, location variability, and sleep metrics. We utilize the data standardization platform for reproducibility provided by the authors~\cite{globem_platform}.

While GLOBEM includes thousands of features, we select just four features to employ, each evaluated daily: smartphone moving/static time ratio, total screen time (minutes), total sleep time (minutes), and total steps. We train Gaussian HMMs on these continuous features, depicted for three anonymous participants in \ref{fig:features_sample}. Our small feature set allows us to learn meaningful correlations and avoid converging to degenerate or redundant models due to insufficient samples. Most deep-learning based models included in the benchmark leverage the 14-day history versions of the provided features, which sum over the prescribed time period. This helps reduce the frequency of missing values, but results in a lagging time series. We use the raw feature for the current day, rather than the 14-day history, which we find boosts performance by $\sim3\%$ for our \hmme approach.
Sequences are constructed from a 28-day history of normalized features, aggregated by participant and preprocessed using the provided data platform. We filter out days where more than half of these features are missing, and within each study participant, fill remaining missing values using median imputation. After filtering, we have 5,393 training samples from 2018, 3,228 from 2019, 2,036 from 2020, and 1,192 from 2020, with class ratios (not-depressed to depressed) of 1.13, 1.23, 1.29, and 1.14 respectively.

\textbf{Evaluation \hspace{0.5em}}
We use the "all-but-one" validation scheme \cite{globem_platform}, training on three years and testing on the fourth. Performance metrics include AUC-ROC and balanced accuracy (average of specificity and sensitivity), which we adopt as in \cite{globem} for its robustness to class imbalance \cite{balanced_acc}. We compare our approach to the top-performing model from the GLOBEM study \cite{globem}, \textit{Reorder}, a CNN-based deep learning algorithm, along with a traditional SVM-based method (Canzian et al. \cite{canzian}) from the benchmark. We also compare against a Random Forest-based approach included in the benchmark, based on Wahle et al. \cite{wahle}. We train the Random Forest approach on our selected four features rather than the original paper's six, using 450 trees, with the number of leaf nodes selected via K-Folds cross validation on a small training subset.

\begin{table*}[tb]
\centering
\begin{tabular}{l c c c c c}
\toprule
 & \textbf{SVM \cite{canzian}} & \textbf{Random Forest \cite{wahle}} & \textbf{CNN (Reorder)} & \textbf{HMM} & \textbf{HMM-e} \\
\midrule
\textit{AUC-ROC} & 49.9 $\pm$ 2.9 & 53.6 $\pm$ 2.1 & 56.3 $\pm$ 0.8 & 51.8 $\pm$ 1.7 & 53.6 $\pm$ 1.3 \\
\textit{Balanced Acc.} & 49.8 $\pm$ 1.2 & 50.7 $\pm$ 0.5 & 54.7 $\pm$ 1.6 & 51.8 $\pm$ 1.7 & 52.5 $\pm$ 0.9 \\
\bottomrule
\end{tabular}
\caption{Balanced Accuracy and AUC-ROC on GLOBEM. Our approach outperforms baseline machine learning methods and achieves similar results to the best-performing deep learning approach.}
\label{table:globem_performance}
\end{table*}

\textbf{Results \& Discussion \hspace{0.5em}} \label{sec:results}
Our approach using just four features outperforms machine learning approaches like \cite{canzian} which uses up to 17 features, and performs comparably to or outperforms many other machine learning approaches included in the benchmark in \cite{globem}. Our mean AUC-ROC and balanced accuracy using singleton HMMs beat out \cite{canzian} by 1.9 and 2 percentage points, respectively. Using \hmme, our AUC-ROC and balanced accuracy beat the same by 3.7 and 2.7 percentage points. In terms of balanced accuracy, \hmme outperforms \cite{wahle} 1.8 percentage points while achieving the same AUC-ROC. We also achieve similar performance as the complex deep-learning approach \textit{Reorder}, falling 2.7 percentage points short in AUC-ROC and 2.2 in balanced accuracy. Notably, we achieve this performance with traditional machine learning techniques, simpler models, and fewer features. 

For each of our $N \times M$ base learners with \texttt{num\_states} states, on \texttt{num\_features} features, we learn $\texttt{num\_states} + (\texttt{num\_states} \times \texttt{num\_states}) + (\texttt{num\_states} \times \texttt{num\_features})$ parameters. In our case, with 4 features and 3 states, this results in 6,000 total parameters -- versus \textit{Reorder}'s 10,099 parameters. On our hardware, detailed in \ref{sec:hardware_software}, \hmme trains on 2019-2021 data in 5 minutes versus 18 minutes for \textit{Reorder}, 3 minutes for \textit{Wahle}, and 29 seconds for \textit{Canzian}. 

In an unsupervised setting, training \hmme without class labels and using subsampled data, we perform clustering with K-Means and dimensionality reduction with UMAP; results shown in Fig. \ref{fig:umap_unsupervised}.

\section{Conclusion}

We explore the connection between human behavior modeling and sequence analysis, providing a general framework for extracting coherent sequences from fragmented data. We present HMM-\emph{e}, an ensemble learning approach that effectively models behavior sequences with minimal feature engineering. Our experiments demonstrate that \hmme outperforms traditional machine learning baselines and delivers results comparable to complex deep-learning models, despite using fewer features. This highlights the efficiency and potential of our approach for scalable and interpretable sequence modeling in behavior-driven applications.

\clearpage
\bibliographystyle{references}
\bibliography{references}

\bigskip
\appendix

\section{Supplementary Material: Extended Results}

\begin{table*}[htb]
\centering
\begin{tabular}{l l c c c c c}
\toprule
\textbf{Method}& & \textbf{2018} & \textbf{2019} & \textbf{2020} & \textbf{2021} & \textbf{Mean} \\
\midrule
\multirow{2}{*}{\textbf{Random Forest} (\textit{Wahle et al.})} & \textit{AUC-ROC}& 51.9 & 51.8 & 55.8 & 54.9 & 53.6 $\pm$ 2.1 \\
                                  & \centering\textit{Balanced Acc.}& 51.2   & 50.0   & 51.1   & 50.7   &  50.8 $\pm$ 0.5  \\ \midrule
\multirow{2}{*}{\textbf{SVM} (\textit{Canzian et al.})}& \textit{AUC-ROC}& 49.1 & 48.4 & 48.0 & 54.2 & 49.9 $\pm$ 2.9 \\
                                  & \centering\textit{Balanced Acc.}& 48.0 & 50.4 & 50.6 & 50.1 & 49.8 $\pm$ 1.2  \\ \midrule
\multirow{2}{*}{\textbf{CNN} (\textit{Reorder})} & \textit{AUC-ROC}& 56.7& 56.4& 55.2& 57.1& 56.3 $\pm$ 0.8 \\
                                  & \centering\textit{Balanced Acc.}& 54.8   & 54.2   & 53.0   & 56.8   &  54.7 $\pm$ 1.6  \\ \midrule
\multirow{2}{*}{\textbf{HMM}}     & \textit{AUC-ROC} &  52.4 &  50.9   & 54.0   & 50.0   & 51.8 $\pm$ 1.7   \\ 
                                  & \textit{Balanced Acc.}& 52.4 &  50.9   & 54.0   & 50.0   & 51.8 $\pm$ 1.7   \\ \midrule
\multirow{2}{*}{\textbf{HMM-e}}   & \textit{AUC-ROC} & 51.9   & 54.7   & 54.4 & 53.3   & 53.6 $\pm$ 1.3   \\ 
                                  & \textit{Balanced Acc.}& 52.2   & 52.3  & 53.8   & 51.8   & 52.5 $\pm$ 0.9\\ 
\bottomrule
\end{tabular}
\caption{Balanced Accuracy and AUC-ROC on GLOBEM data. For each year indicated, we train on the other 3 years, and hold out that year for testing.}
\label{table:globem_performance_extended}
\end{table*}

In addition to the aggregated results in Table \ref{table:globem_performance}, we present results for each of the four years in the GLOBEM dataset in Table \ref{table:globem_performance_extended}. Each year indicated is the held-out test year, while the other three are used for training. We compare our approaches against the best-overall-performing approach in the GLOBEM study, \textit{Reorder} \cite{globem}. \textit{Reorder} is a deep-learning based approach tailored for the GLOBEM problem, built on a CNN backbone.  

We also present results from \cite{canzian}, an SVM-based traditional machine learning approach. We find that we are consistently able to outperform \cite{canzian}, and in terms of balanced accuracy, achieve comparable results to \textit{Reorder} with far fewer features and much less complex models.

\begin{figure*}[b]
\centering
\includegraphics[width=1.0\linewidth]{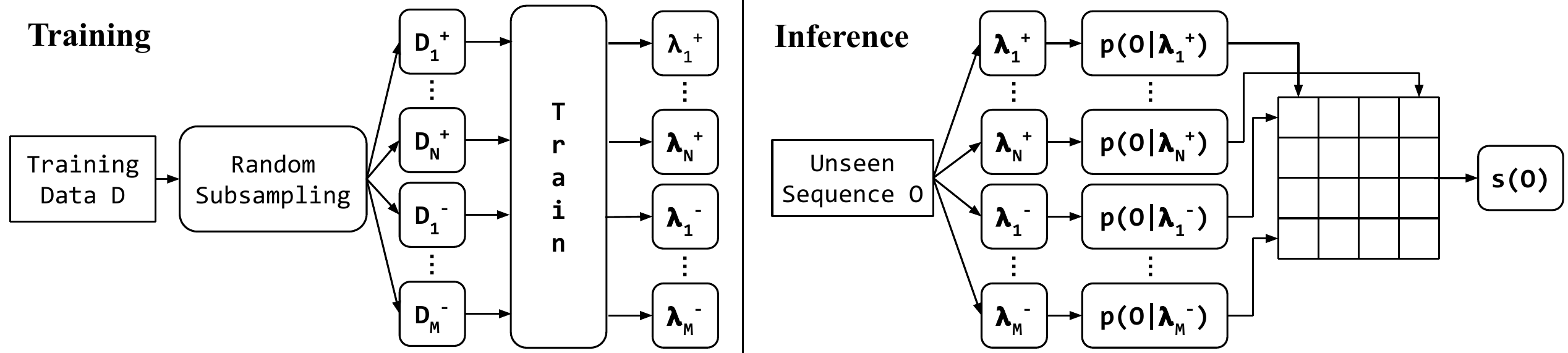} 
\caption{Flow diagram of our \hmme ensemble training and inference approach, as detailed in Section \ref{sec:hmm_ensembles}. While we adopt this approach using HMMs, the framework itself is model agnostic. The training data is broken into random subsets, and a diverse ensemble of learners is trained on these subsets. At inference time, pairwise matchups of likelihoods given by the models are compared, giving the composite score $s$.}
\label{fig:approach_diagram_extended}
\end{figure*}

We illustrate the training and inference pipelines of the \hmme approach detailed in Section \ref{sec:formalization}, in Figure \ref{fig:approach_diagram_extended}. 

\begin{figure*}[htb]
\centering
\includegraphics[width=1.0\linewidth]{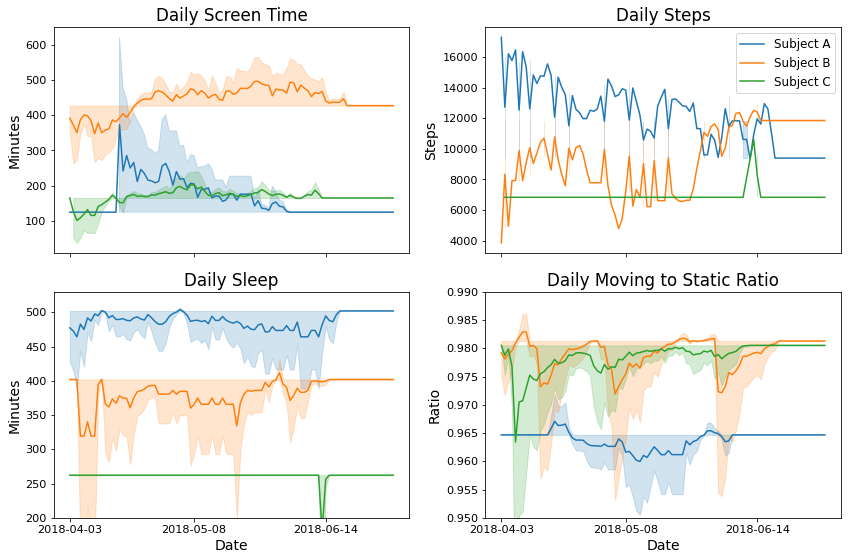}
\caption{Samples of the daily features we model, across 3 anonymized 2018 participants. For visualization purposes, features are lightly smoothed using an exponential weighted moving average with a half-life of 4 days. For modeling, features are normalized.}
\label{fig:features_sample}
\end{figure*}

\section{Feature Selection}

Figure \ref{fig:features_sample} shows the four features we choose to model, across three anonymized participants from the 2018 study. While many baseline models included in the benchmark choose to model tens or even hundreds of features, we use a small feature set. This allows us to learn meaningful correlations and avoid converging to degenerate models due to insufficient samples. We include two smartphone features (screen time and location-based moving-to-static ratio) and two wearable features (sleep time and steps). All of these features are computed daily.

\section{Hardware and Software Stack} \label{sec:hardware_software}

Our experiments are performed on an AWS \textit{r5.24xlarge} EC2 instance featuring 96 virtual CPUs and 768 GB of memory. Due to the lightweight nature of the models trained, we do not have to leverage GPU acceleration. The environment is configured with Ubuntu 20.04 LTS as the operating system, and we use Python version 3.8.10. Aside from standard machine-learning libraries like Pandas, NumPy, Pytorch, Scikit-Learn, and Tensorflow, we also use HMMLearn to train our HMMs, and Ray to parallelize training and data processing.

\end{document}